\documentclass[acmtog]{acmart}

\usepackage{bm}

\definecolor{Highlight}{HTML}{39b54a}  

\usepackage{pifont}
\usepackage{algorithm}
\usepackage{listings}

\newif\ifdraft
\drafttrue
\ifdraft
\newcommand{\gqc}[1]{{\color{orange}[\textbf{Gordon:} #1]}}
\newcommand{\kac}[1]{{\color{purple}[\textbf{Kfir:} #1]}}
\newcommand{\dcc}[1]{{\color{red}[\textbf{Danny:} #1]}}
\newcommand{\jacksoncomment}[1]{{\color{blue}[\textbf{Jackson:} #1]}}
\newcommand{\opc}[1]{{\color{teal}[\textbf{Or:} #1]}}
\newcommand{\dosc}[1]{{\color{magenta}[\textbf{Daniil:} #1]}}

\newcommand{\todo}[1]{{\textbf{\color{red}[TODO: #1]}}}

\else
\newcommand{\gqc}[1]{}
\newcommand{\kac}[1]{}
\newcommand{\dcc}[1]{}
\newcommand{\jacksoncomment}[1]{}
\newcommand{\opc}[1]{}
\newcommand{\todo}[1]{}

\newcommand{\dosc}[1]{}
\fi
\usepackage{etoolbox}
\makeatletter
\AfterEndEnvironment{algorithm}{\let\@algcomment\relax}
\AtEndEnvironment{algorithm}{\kern2pt\hrule\relax\vskip3pt\@algcomment}
\let\@algcomment\relax
\newcommand\algcomment[1]{\def\@algcomment{\footnotesize#1}}
\renewcommand\fs@ruled{\def\@fs@cfont{\bfseries}\let\@fs@capt\floatc@ruled
  \def\@fs@pre{\hrule height.8pt depth0pt \kern2pt}%
  \def\@fs@post{}%
  \def\@fs@mid{\kern2pt\hrule\kern2pt}%
  \let\@fs@iftopcapt\iftrue}
\makeatother
\newcommand{\cmmnt}[1]{}

\usepackage{enumitem}

\AtEndPreamble{
    \usepackage[capitalize]{cleveref}
    \crefname{section}{Sec.}{Secs.}
    \Crefname{section}{Section}{Sections}
    \Crefname{table}{Table}{Tables}
    \crefname{table}{Tab.}{Tabs.}
}

\AtBeginDocument{%
  }

\setcopyright{acmlicensed}
\copyrightyear{2025}
\acmYear{2025}
\acmConference[Preprint]{Preprint}{September}{2025}
\acmISBN{978-1-4503-XXXX-X/2018/06}

\newcommand{\methodname}{ComposeMe}

\newcommand{\newtext}[1]{#1}

\acmSubmissionID{2203}

\renewcommand\footnotetextcopyrightpermission[1]{}

\usepackage{hyperref}
\usepackage{xcolor}
\definecolor{pink}{rgb}{1.0, 0.6, 0.9} 
\hypersetup{
    colorlinks=true,
    urlcolor=pink, 
}
\newcommand{\webpage}{\href{https://snap-research.github.io/composeme/}{\textcolor{pink}{https://snap-research.github.io/composeme/}}}

\begin{document}

\title{ComposeMe: Attribute-Specific Image Prompts for Controllable Human Image Generation}
\author{Guocheng Gordon Qian}
\orcid{https://orcid.org/0000-0002-2935-8570}
\author{Daniil Ostashev}
\orcid{https://orcid.org/0009-0009-7259-7399}
\author{Egor Nemchinov}
\orcid{https://orcid.org/0009-0007-9492-6574}
\author{Avihay Assouline}
\orcid{https://orcid.org/0009-0009-2126-4324}
\author{Sergey Tulyakov}
\orcid{https://orcid.org/0000-0003-3465-1592}
\author{Kuan-Chieh Jackson Wang}
\orcid{https://orcid.org/0000-0002-6785-8146}
\author{Kfir Aberman}
\orcid{https://orcid.org/0000-0002-4958-601X}
\affiliation{%
  \institution{Snap Inc.}
  \state{Canifornia}
  \country{USA}
}

\renewcommand{\shortauthors}{Qian et al.}

\begin{abstract}
Generating high-fidelity images of humans with fine-grained control over attributes such as hairstyle and clothing remains a core challenge in personalized text-to-image synthesis. While prior methods emphasize identity preservation from a reference image, they lack modularity and fail to provide disentangled control over specific visual attributes.
We introduce a new paradigm for attribute-specific image prompting, in which distinct sets of reference images are used to guide the generation of individual aspects of human appearance, such as hair, clothing, and identity. Our method encodes these inputs into attribute-specific tokens, which are injected into a pre-trained text-to-image diffusion model. This enables compositional and disentangled control over multiple visual factors, even across multiple people within a single image.
To promote natural composition and robust disentanglement, we curate a cross-reference training dataset featuring subjects in diverse poses and expressions, and propose a multi-attribute cross-reference training strategy that encourages the model to generate faithful outputs from misaligned attribute inputs while adhering to both identity and textual conditioning.
Extensive experiments show that our method achieves state-of-the-art performance in accurately following both visual and textual prompts. Our framework paves the way for more configurable human image synthesis by combining visual prompting with text-driven generation. Our project page is available at {\webpage}. 

\end{abstract}


\begin{teaserfigure}
  \includegraphics[width=\textwidth]{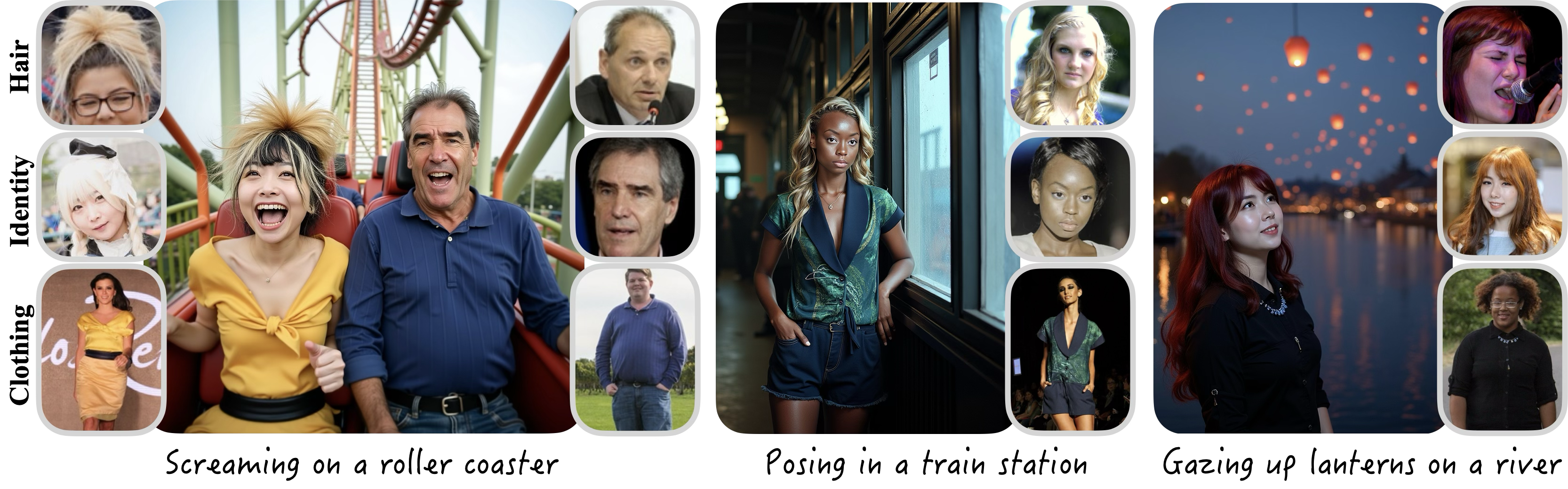}
  \caption{Given separate, attribute-specific image prompts for hairstyle, identity, and clothing, our method composes high-fidelity human images with fine-grained control over each attribute. Our results show how different attributes can be recombined to generate diverse, photorealistic outputs with disentangled control.}
  \label{fig:teaser}
\end{teaserfigure}


\maketitle

\section{Introduction}

The advent of large-scale text-to-image (T2I) diffusion models \cite{rombach2022high, saharia2022photorealistic} has revolutionized the field of generative AI, enabling the synthesis of diverse and high-quality images from textual descriptions.  A pivotal extension of this technology is personalization \cite{textual_inversion, ruiz2023dreambooth}, which empowers users to incorporate their own specific subjects into the generation process and significantly expands the creative possibilities of T2I models.

Despite these advancements, achieving fine-grained injection of an identity presents ongoing challenges. There is a clear need for a framework that allows users to provide distinct visual references for different attributes of an identity and compose them seamlessly and faithfully. However, existing methods are restricted to considering a whole identity as a single concept~\cite{ruiz2023dreambooth, ye2023ip-adapter, Omni-ID, po2023orthogonal, wang2024moa, patashnik2025nested}. 
Simultaneously controlling specific, interchangeable visual attributes of a single identity, such as a particular face, a distinct hairstyle from another source, and specific clothing items from a third source, remains largely unexplored and difficult to achieve with current techniques. 
A compositional approach to identity enables a new class of applications that demand modularity and user-driven customization. 
These include a more freeform virtual try-on systems where users can preview how different hairstyles or outfits look on a given face in diverse contexts specified by a text prompt, or creative tools for designers to explore novel combinations of visual traits across personas~\cite{han2018viton,kim2024stableviton,xu2025ootdiffusion}. 
It also opens the door to personalization pipelines that are more flexible and reusable, allowing each attribute to be curated, updated, or swapped independently without retraining the entire identity representation.

To address this gap, we introduce \emph{Attribute-Specific Image Prompts}, a novel framework for controllable human image generation that specializes in composition of identity attributes. Attribute-Specific Image Prompts allows users to synthesize images by specifying the distinct identity attributes such as facial identity, hairstyle, and clothing through separate sets of reference images. As illustrated in ~\cref{fig:teaser}, a diffusion model with Attribute-Specific Image Prompts can generate an harmonized image of persons by taking visual attributes from different sources and guided by a text prompt. 
With high fidelity to each input attribute, our approach can synthesize a coherent image in a fully imagined scene, seamlessly combining a given facial identity with the desired hairstyle and outfit. Furthermore, our method extends naturally to multi-subject generation—a particularly challenging scenario in personalized image synthesis—enabling the composition of multiple distinct identities within a single frame.

The core technical novelty of our work lies in two fold. First, we propose \emph{\methodname{}}, a model that employs Attribute-Specific Image Prompts by using dedicated tokenizers to process reference images for each distinct visual attribute separately. These attribute-specific tokens are simply merged and then injected into a frozen pre-trained text-to-image diffusion model. 
Our approach offers \emph{fine-grained control from multiple image sets}, allowing for a mix-and-match capability for visual attributes that surpasses the flexibility of many existing personalization techniques. 
Second, we propose \emph{Multi-Attribute Cross-Reference Training}. Without this training strategy, certain attributes—such as clothing, face, and hair—remain inherently entangled with confounding factors like body pose, expression, or head orientation in single-reference training setups. As a result, naively combining image prompts from different sources can lead to unnatural, collage-like artifacts that compromise visual coherence and reduce adherence to text prompts. Our multi-attribute cross-reference training explicitly decouples these entanglements by supervising generation with inputs and targets sourced from different individuals and poses, enabling the model to synthesize harmonized outputs that maintain fidelity across composited attributes. This training formulation is key to avoiding undesirable copy-paste effects and achieving natural, seamless integration of independently sourced visual traits. We list our main contributions as follows:
\begin{itemize}[topsep=0pt, partopsep=0pt, leftmargin=1.5em]
    \item We introduce \methodname{}, a novel method for multi-attribute personalized image generation that enables the composition of facial identity, hairstyle, and clothing from distinct image inputs.
    \item We propose an attribute-specific tokenization strategy, where dedicated tokenizers process reference images for each visual component including face, hair, and clothing to capture attribute-specific features. This human-centric design achieves significantly higher fidelity to the input attributes compared to generic approaches such as GPT-4o.
    \item We devise a novel adapter training strategy, \emph{multi-attribute cross-reference}, in which the inputs are intentionally misaligned—both with each other and with the target—to facilitate disentangled appearance injection. 
    Through extensive experiments, we demonstrate that this approach leads to more coherent composition and higher image quality.
    \item We validate the effectiveness of \methodname{} across various personalization settings, including the challenging Multi-Attribute Multi-ID setting, for which no prior method provides a solution.
\end{itemize}

\begin{figure}[t]
\centering
\includegraphics[width=\linewidth]{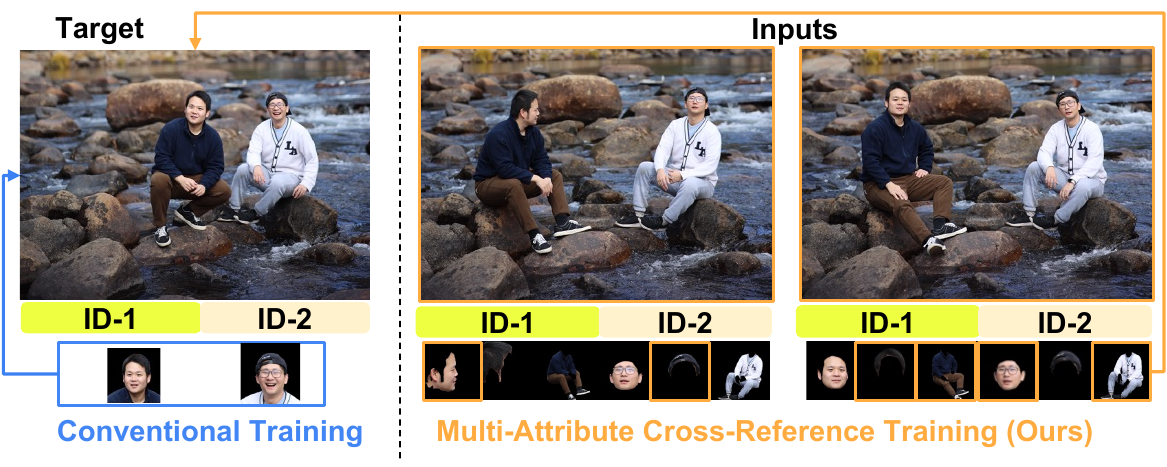}
\caption{\newtext{\textbf{Conventional single-reference adapter training (left)} treats an identity with multiple attributes as a single, indivisible object, cropped directly from the target. In contrast, our \textbf{Multi-Attribute Cross-Reference Training (right)} is attribute-aware: it decomposes identity into distinct visual attributes (face, hair, clothing), sourcing each from different inputs and predicting separate image as target. Our approach curates pairs of misaligned inputs and natural target, therefore enabling generation of coherent outputs even from misaligned attribute inputs in inference.
}}
\label{fig:cross-reference-data}
\end{figure}

\section{Related Work}

\noindent\textbf{Personalized Generation.} Pioneering work Textual Inversion \cite{textual_inversion} learns a new token embedding in the text tokenizer's space to represent a novel concept. This method is convenient, but cannot capture the fine-grained details of the subject. Approaches like MyStyle~\cite{nitzan2022mystyle} and DreamBooth~\cite{ruiz2023dreambooth} are capable of high-fidelity subject injection by fine-tuning parts of the GAN~\cite{goodfellow2020gan, gal2021stylegan} or diffusion backbone~\cite{ho2020denoising,ldm}, but are computation intensive. 
More recent efforts~\cite{ye2023ip-adapter, wang2024moa, wang2024instantid, fastcomposer, gal2024lcm, patashnik2025nested} focus on lightweight adapters that enable personalization with frozen base model, offering greater efficiency and flexibility.
IP-Adapter~\cite{ye2023ip-adapter} is a seminal work in this category, introducing an image prompt adapter that uses a decoupled cross-attention mechanism to inject image features into a frozen T2I model. 
Follow-up works InstantID~\cite{wang2024instantid}, InfiniteYou~\cite{InfiniteYou}, LCM-LookAhead \cite{gal2024lcm}, PuLID~\cite{PuLID}, and ID-Aligner~\cite{ID-Aligner} improve identity preservation by employing identity modules, ID losses and reward functions. Omni-ID~\cite{Omni-ID} improves facial representation tuned for generative tasks with a few-to-many training objective.
Our work differs from prior approaches by introducing attribute-specific image prompts, allowing each aspect of a person's appearance—such as face, hairstyle, and clothing—to be specified using separate sets of reference images. This departs from the conventional one-image-per-identity paradigm and enables finer-grained, modular control. 

\noindent\textbf{Multi-Concept Personalization.}
Achieving high-fidelity concept injection while generating harmonized images—where each visual segment is reposed appropriately according to the prompt—remains a significant challenge. Optmization-based methods typically focus on learning disentangled representations of concepts~\cite{custom-diffusion,avrahami2023break,garibi2025tokenverse} or training multiple LoRAs~\cite{hu2022lora} to represent input concepts~\cite{po2023orthogonal,kong2024omg,gu2024mix,yang2024lora,abdal2025dynamic}. However, these approaches are computationally expensive and become prohibitive when scaling to a large number of users.
In contrast, optimization-free methods have gained popularity due to their efficiency and versatility. These methods commonly rely on image tokenizers and adapters, often coupled with complex module designs and regularization losses, to enable feed-forward personalization~\cite{parmar2025object,patel2024lambda,xiao2024fastcomposer,li2024omniprism,wang2024moa,han2024emma, he2024uniportrait, MIP-Adapter, MS-Diffusion}. However, they are generally limited to treating identity as a single visual unit and lack the capability to compose multiple distinct visual prompts for single identity. 
Recent works~\cite{pan2023kosmos,xie2024show,zhou2024transfusion, mi2025thinkdiff} such as OmniGen~\cite{OmniGen} \newtext{and UNO~\cite{UNO}} have begun exploring more general compositional generation. Nonetheless, these generalists do not focus on human-centric personalization and struggle to maintain high-fidelity synthesis. In contrast, our approach is specifically tailored for human image generation, achieving significantly better performance.

\noindent\newtext{\textbf{Controllable Portrait Generation} — Unlike personalization methods that place humans in various poses and contexts via text prompts, another relevant line of research focuses on controllable portrait generation. A typical example is virtual try-on \cite{pix2pixhd}, where, given a source photo of a person and an image of clothing, the method replaces the clothing in the photo with the specified garment. Recent approaches \cite{parts2whole, BootCamp, MIP-Adapter, MS-Diffusion} leverage diffusion models to extend controllable generation beyond clothing, incorporating elements such as poses, objects, and garments. These methods are generally limited to portraits and require pre-aligned product garment images, whereas our approach handles arbitrary clothing, hairstyles, and faces from misaligned, real-world photos.}

\section{Methodology}
\label{sec:method}

\begin{figure}[t]
\centering
\includegraphics[width=1.0\columnwidth]{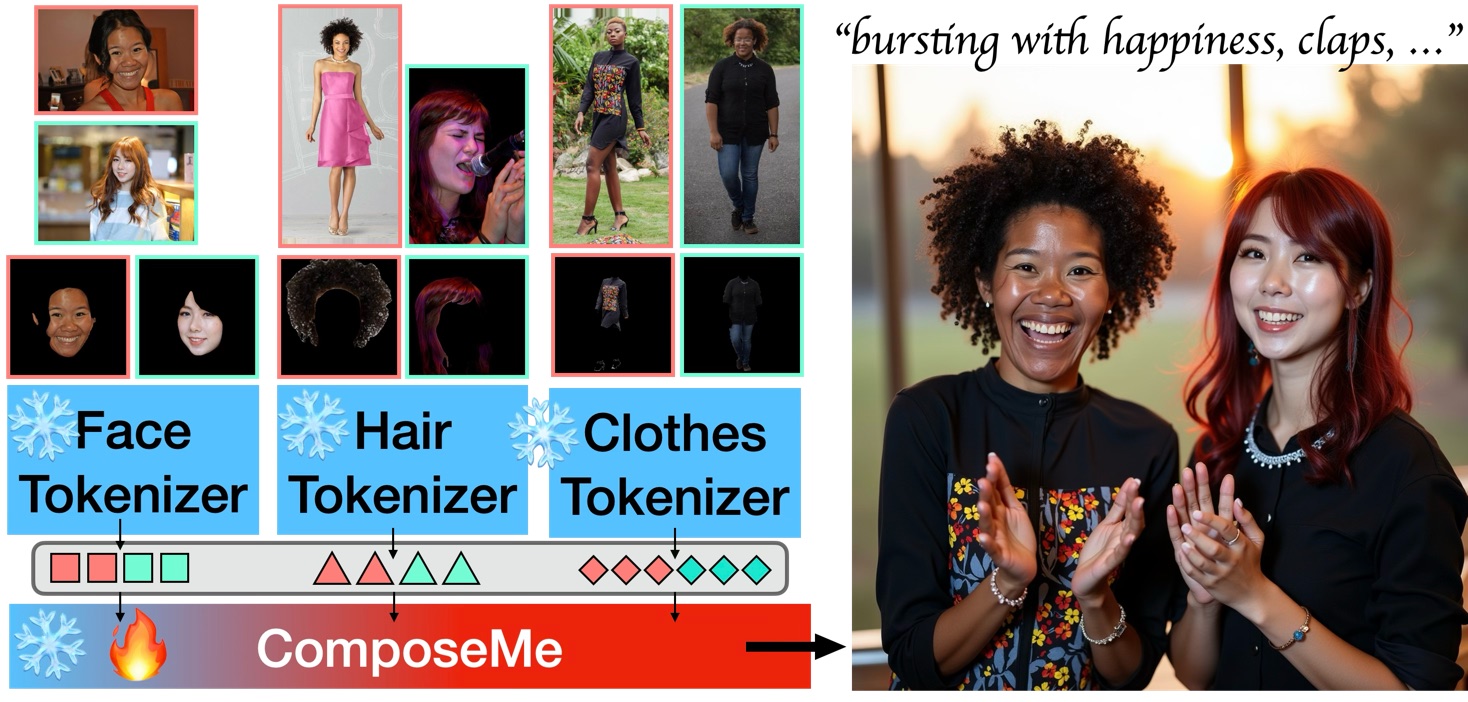} 
\caption{\textbf{Inference pipeline of \methodname}. Attribute-Specific Image Prompts are processed by specific tokenizers. The merged tokens are injected into a pre-trained diffusion model.}
\label{fig:architecture}
\end{figure}

This work introduces \textit{\methodname} that is conditioned the proposed \textit{Attribute-Specific Image Prompts}, which aims to personalize multiple visual attributes in text-to-image generation. 
Unlike previous arts that personalize whole objects through a single input image, \methodname~decomposes each human subject into configurable \textit{attributes}: identity (face), hairstyle, and clothing. Distinct image sets are leveraged for each attribute across different subjects, allowing for a versatile capture of visual attributes. 
Through attribute-specific tokenizer, token merging, and shared decoupled cross-attention, \methodname~injects this multi-ID multi-attribute information into a pre-trained diffusion model in an efficient manner. 
A key challenge in this task lies in the misalignment of multiple attributes during inference. For instance, the hair contour may not match the head shape, or the composition of multiple visual prompts may conflict with the intended semantics of the text prompt. To address this, we propose \textit{Multi-Attribute Cross-Reference} training, that uses different source image for each attribute as input to generate the target image. As a result, \methodname~outputs naturally aligned outputs from misaligned input image sets. 

\subsection{ Attribute-Specific Image Prompts}
\label{ssec:composable_person}

We introduce Attribute-Specific Image Prompts that decomposes an subject into several attributes, each can be represented by a distinct set of reference images. Mathematically speaking, given subject $I$, Attribute-Specific Image Prompts are represented by: 
\begin{equation}
    I = \{I_{p_0}, \dots, I_{a_i}, \dots, I_{p_{K-1}}\}, i \in (0, K-1)
\end{equation}
where $K$ is the number of visual attributes $p$ that $I$ is composed of. Refer to \cref{fig:architecture} for a specific example, where $K=3$ and $a_i$ denotes face, hairstyle, clothing, individually for $i=0, 1, 2$. 

Attribute-Specific Image Prompts enables new application by combining different visual attributes into one generation, where each attribute $a_i$ may come from a distinct source image. 

\subsection{\methodname~Pipeline}
\label{ssec:omni_adapter_arch}
We present an adapter based solution for using Attribute-Specific Image Prompts in personalized generation, namely \methodname. As illustrated in \cref{fig:architecture}, our approach involves three stages: (1) Attribute-specific tokenization, (2) Multi-attribute merging, and (3) Injection into a pre-trained diffusion model.

\subsubsection{Attribute-Specific Tokenization}
\label{sssec:attribute_specific_encoders}
To capture unique information from each visual component, \methodname~employs dedicated feature tokenization pipelines. Each attribute uses a distinct set of images to define a different visual attribute of the subject. Each attribute receives a set of images as input denoted as $I_{a_i}$ which are tokenized as $L_{a_i}$ through a dedicated tokenizer $E_{a_i}$ and a linear projection layer $W_{a_i}$ to align the dimensions across  all $E_a$. The subscript $a$ stands for `attribute' and $i$ represents the $i$-th attribute. The aforementioned process is mathematically given as:
\begin{equation}
    L_{a_i} = W_{a_i}\left(E_{a_i}(I_{a_i})\right)
\end{equation}
where $I_{a_i} = \{I_{a_i}^0, I_{a_i}^1, \dots, I_{a_i}^{{N_i}-1}\}$ denotes the inputs for attribute $a_i$, each input $I_{a_i}^j \in \mathbb{R}^{3\times H \times W}$ represents an input RGB image, and $N_i$ denotes the number of images used in $a_i$. $E_{a_i}$ tokenizes $a_i$.  A linear layer $W_{a_i} \in \mathbb{R}^{{D_{a_i}}\times{D}}$ projects the tokens of $a_i$ from dim $D_{a_i}$ to the same dimension $D$ for all attributes.

The above tokenization is applied to each attribute separately. Each attribute uses their dedicated tokenizer as illustrated in \cref{fig:architecture} and studied in \cref{subsec:ablation_encoders}. This attribute-specific tokenization stage returns distinct tokens for each attribute: ${L_{a_0}, L_{a_1}, \dots, L_{a_K}}$.

\subsubsection{Multi-Attribute Merging}
\label{sssec:merging_strategy}
\methodname~merges $K$ attribute tokens to form the multi-attribute subject representation $e$ as next:
\begin{equation}
{L_e} = [L_{p_0} + \mathbf{PE}_a^{0}; L_{p_1} + \mathbf{PE}_a^{1}; \dots; L_{p_K} + \mathbf{PE}_a^{K}],
\label{eq:merge_attributes}
\end{equation}
where $\mathbf{PE}_a$ is a separate learnable positional embedding added to all tokens of each attribute before merging to distinguish attributes.  
\methodname~uses the simple token-wise concatenation as the merging strategy. The length of ${L_e}$ equals $\sum_{0}^{K-1} |L_{a_i}|$. When a certain attribute is not provided, the unconditional embedding of $L_{a_i}$ computed on a black RGB (i.e. all zeros) image is used. 

To support more than $1$ subject in the generation, \methodname~allows merging the representation across subjects to form the \emph{final multi-attribute representation $L$} as follows:
\begin{equation}
{L} = [L_{e_0} + \mathbf{PE}_e^{0}; L_{e_1} + \mathbf{PE}_e^{1}; \dots; L_{e_M} + \mathbf{PE}_e^{M}],
\label{eq:merge_subject}
\end{equation}
where $L_{e_j}$ is the representation for subject $j$ and $M$ denotes the maximum number of subjects to support, and $\mathbf{PE}_e$ is a positional embedding for the subject. 
Experiments show that this simple concatenation with the necessary positional embeddings produces satisfactory results.

\subsubsection{Adapter Injection}
\label{sssec:token_injection}

\methodname~uses the decoupled cross attention \cite{ye2023ip-adapter} to inject multi-attribute representation $L$ into a pre-trained diffusion model. 
Let $Q$ be the query features from an intermediate attention layer in the diffusion model. The original cross-attention computes $Attn_T = \text{CrossAttn}(Q, K_T, V_T)$, where $K_T$ and $V_T$ are derived from text embedding $T$.
A new module computes $Attn_{L}$ computed by multi-attribute representation $L$ is obtained by a new set of learnable decoupled attentions $Attn_L = \text{CrossAttn}_\theta(Q, K_L, V_L)$. $K_L, V_L$ is projected from $L$ by a shared learnable projection model across timestep: $K_L, V_L = W_L^K(L), W_L^V(L)$.
The outputs from frozen prior modules and new decoupled modules are combined by a weighted addition:
\begin{equation}
\mathbf{Z}_{out} = Attn_T + \lambda \cdot Attn_{L}
\label{eq:decoupled_attn}
\end{equation}
where $\lambda \in (0, 1)$ is a scaling factor that balances the influence of the multi-attribute representation from our Attribute-Specific Image Prompts against the text prompt in the inference time. $\lambda$ is set to $1$ during training.





\begin{figure}[t]
\centering
\includegraphics[width=1.0\columnwidth]{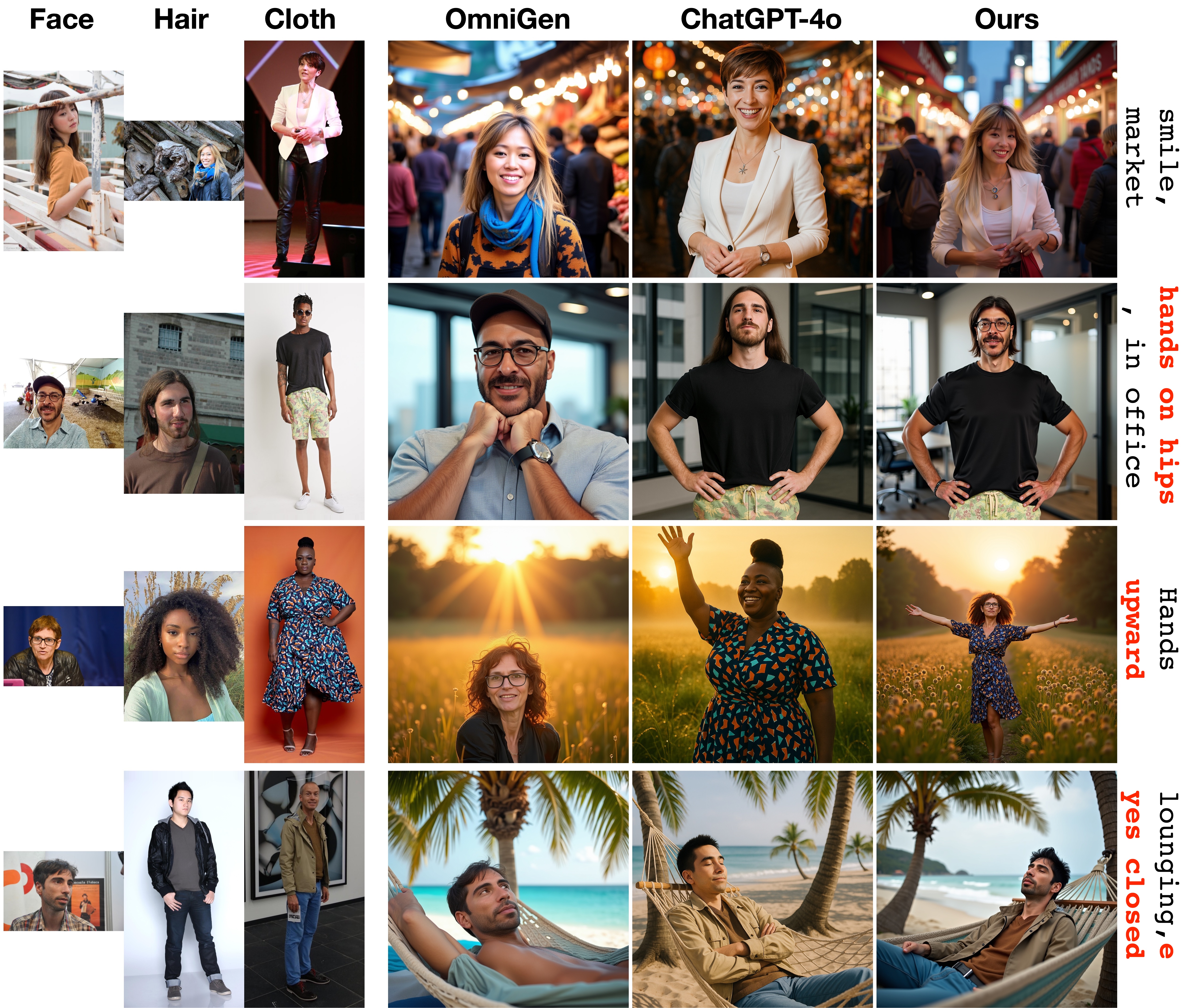}
\caption{\textbf{Qualitative Comparisons in Multi-Attribute, Single-ID Personalization.} Compared to OmniGen and GPT-4o, \methodname~demonstrates significantly improved preservation of ID, hairstyle, and clothing.}
\label{fig:result:multi_attribute}
\end{figure}

\subsection{Multi-Attribute Cross-Reference Training}\label{ssec:training}
We propose a two-stage training strategy: copy-paste pre-training, followed by a novel multi-attribute cross-reference finetuning to optimize a Attribute-Specific Image Prompts system. 

\subsubsection{Multi-Attribute Copy-Paste pre-training}\label{sssec:copy-paste}
This stage trains the \methodname~in a single-reference human image dataset. The input and target are derived from the same image. The flow matching loss~\cite{EsserKSD3} is used in the training:
\begin{equation}
\begin{aligned}
\mathcal{L} &= \mathbb{E}_{x_0 \sim X_0, x_1 \sim X_1, t, T, L} \left[ |
| v - v_\theta(x_t, t, T, L) ||_2^2 \right]
\\
x_t &= x_0 + t*v \\
L &= E(x_1)
\label{eq:loss}
\end{aligned}
\end{equation}
where $v = x_1 - x_0$ denotes the velocity. Due to the input and target ($x_1$) being the same, the multi-attribute representation $L$ is derived from the target $x_1$, i.e. $L=E(x_1)$. \methodname~after this stage is able to inject multi-attribute representation $L$ into the generation, but will copy paste the inputs to the target with minimal ability in changing the visual inputs by text prompts such as updating head pose, expression, and body gesture, and results in unnatural composition  in the generated image.  See \cref{fig:ablation:cross_ref} for visual examples after this stage. 
Despite the undesired copy-pasting result, this stage serves as pre-training that warms up the adapter. A novel multi-attribute cross-reference training stage is proposed as detailed next.

\subsubsection{Multi-attribute Cross-Reference Training}\label{sssec:cross-reference}
This stage trains the \methodname~using the same loss function as \cref{{eq:loss}} but in a multi-reference image dataset. 
This dataset has multiple images $X=\{X_0, X_1, \dots, X_{N-1}\}$, covering diverse expressions, poses, and contexts, for each subject. 
The proposed multi-attribute cross-reference training is formally given as following: 
\begin{equation}
\begin{aligned}
s &\sim U(|X|) \\
x_1 &= X_s \\ 
L &= E(X \setminus X_s)
\label{eq:mult_attribute_loss}
\end{aligned}
\end{equation}

\begin{table}[t]
\centering
\resizebox{\columnwidth}{!}{%
\begin{tabular}{lccccc}
\toprule
\textbf{Method} & \textbf{FaceNet}~$\uparrow$ & \textbf{CLIP-Clothing}~$\uparrow$ & \textbf{CLIP-Hair}~$\uparrow$ & \textbf{CLIP-T}~$\uparrow$ & \textbf{HPSv2 Score}~$\uparrow$ \\
\midrule
\multicolumn{6}{c}{{Multi-Attribute – Single Subject}} \\
\textbf{Ours} & \textbf{0.6870} & \textbf{0.8526} & \textbf{0.8778} & 0.2843 & 0.2904 \\
OmniGen & 0.5496 & 0.7817 & 0.8450 & 0.2893 & \textbf{0.3001} \\
GPT-4o & 0.2992 & 0.8288 & 0.8430 & \textbf{0.2974} & 0.2967 \\
\midrule
\multicolumn{6}{c}{{Single Attribute – Single Subject}} \\
\textbf{Ours} & 0.6881 & -- & 0.8735 & \textbf{0.2877} & \textbf{0.3006} \\
PuLID & 0.7639 & -- & 0.8720 & 0.2799 & 0.2783 \\
Omni-ID & \textbf{0.7873} & -- & \textbf{0.8898} & 0.2688 & 0.2674 \\
InfiniteYou & 0.7788 & -- & 0.8252 & 0.2909 & 0.2947 \\
\midrule
\multicolumn{6}{c}{{Single Attribute – Single Subject – Full Body}} \\
\textbf{Ours} & \textbf{0.6845} & \textbf{0.8631} & \textbf{0.8745} & 0.2820 & 0.2888 \\
OmniGen & 0.6403 & 0.7980 & 0.8522 & 0.2900 & \textbf{0.3057} \\
GPT-4o & 0.6095 & 0.8355 & 0.8618 & \textbf{0.2921} & 0.2976 \\
InstantCharacter & 0.3301 & 0.8392 & 0.8322 & 0.2834 & 0.2856 \\
\midrule
\multicolumn{6}{c}{{Single Attribute – Multi Subject}} \\
\textbf{Ours} & 0.7161 & -- & 0.8756 & \textbf{0.2821} & 0.3110 \\
OmniGen & 0.5650 & -- & 0.8556 & 0.2820 & \textbf{0.3147} \\
StoryMaker & \textbf{0.7453} & -- & 0.8692 & 0.2579 & 0.2283 \\
UniPortrait & 0.6555 & -- & 0.8872 & 0.2560 & 0.2552 \\
MIP-Adapter & 0.3062 & -- & 0.8529 & 0.2778 & 0.2670 \\
MS-Diffusion & 0.5078 & -- & \textbf{0.8932} & 0.2463 & 0.2314 \\
UNO & 0.1799 & -- & 0.8317 & 0.2774 & 0.2997 \\
\bottomrule
\end{tabular}%
}
\caption{Quantitative comparison of different methods across various task settings. We use FaceNet similarity between the input and generated faces to measure identity preservation, \newtext{CLIP feature distance between the segmented and center cropped hair and clothing to measure attribute preservation}, HPSv2 to assess generation quality, and CLIP-T for prompt adherence. Our \methodname{} performs on par with OmniGen and GPT-4o in terms of generation quality and prompt following, while achieving the highest identity preservation across most settings—except for the single-attribute single-subject personalization task, where prior methods tend to generate faces with consistent expressions that the FaceNet model favors.}
\label{tab:numbers}
\end{table}

The input ($x_1$) is randomly sampled from the list of input images of $X$. The other images are randomly sampled as attribute sources, where each attribute $a_i$ is sourced from a different input. 
This input-target setup brings in misalignment in both the attribute level and the image level. 
With the first stage warmup, this multi-attribute cross-reference training stage is encouraged to focus on generating naturally aligned output from misaligned inputs. 
The samples of the two stage training is visualized in \cref{fig:cross-reference-data}. 



\section{Experiments}


\subsection{Multi-Attribute Multi-ID Personalization}\label{ssec:multi_attr_multi_id}
\begin{figure*}[t]
\centering
\includegraphics[width=1.0\textwidth, trim=20 0 0 0, clip]{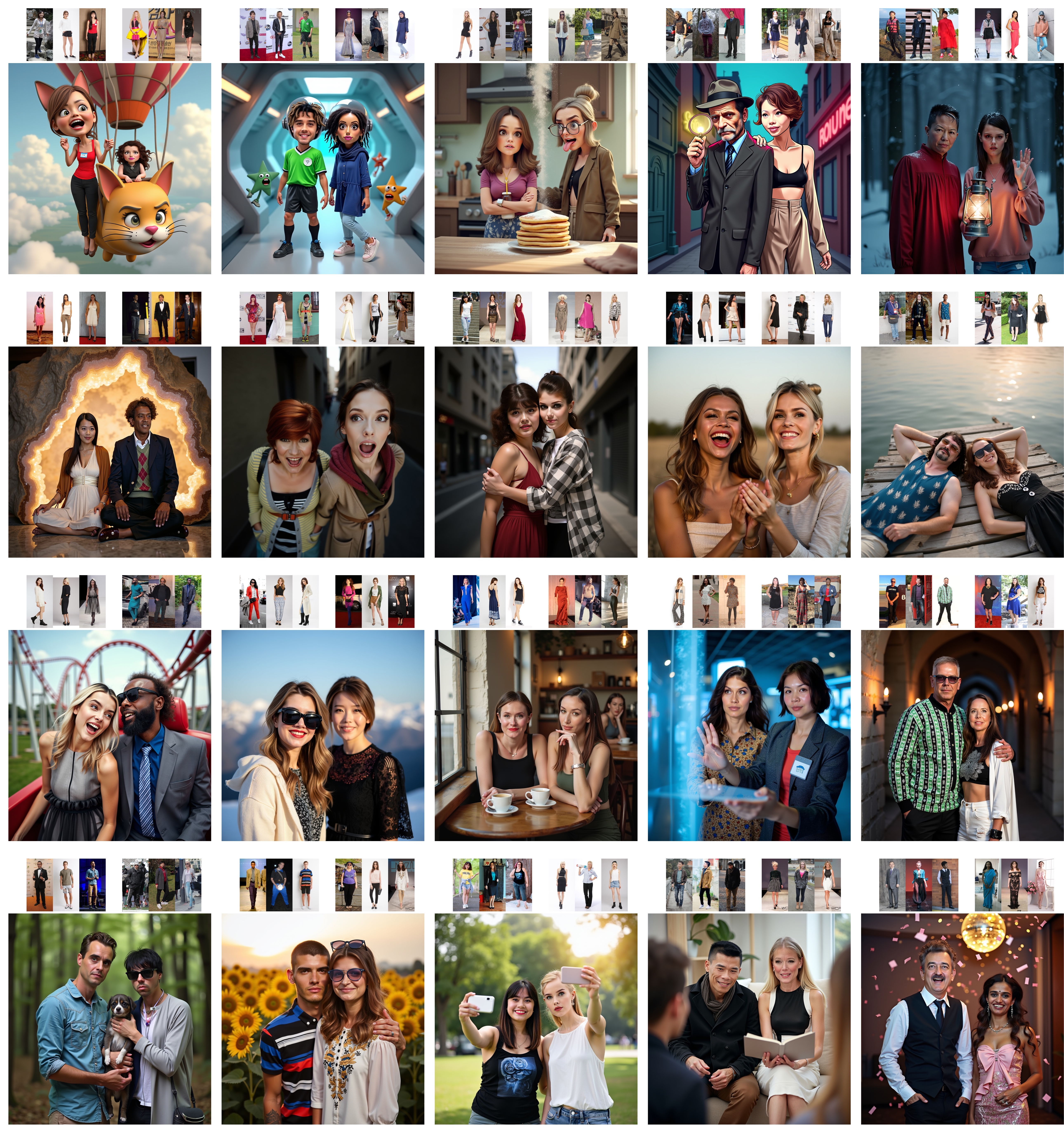}
\vspace{-1.5em}
\caption{\textbf{Gallery of \methodname{} in the task of Multi-Attribute Multi-Identity Personalization.} We show that \methodname{} is capable of generating diverse and high-quality images across a range of scenarios including realistic and stylized generations, preserving identity while adhering to text prompts. 
This capacity of supporting multi-attribute, multi-id generation, where each identity is defined by distinct attributes has not been demonstrated in prior work, to the best of our knowledge. All input images are sourced randomly from SSHQ dataset \cite{StyleGAN-Human}.}
\label{fig:gallery}
\end{figure*}

\methodname~naturally supports multi-attribute, multi-subject generation, where each subject is defined by \textit{three distinct attribute-specific image prompts}: a face image, a hairstyle image, and a clothing image. To the best of our knowledge, this capability has not been demonstrated in prior work. We showcase our results in Fig. \ref{fig:teaser},\ref{fig:architecture}  and more in \cref{fig:gallery}. As observed, our approach is able to generate high-fidelity images across different styles (realistic, cartoon, 3D, etc.) from multiple misaligned image sources. The generation effectively preserves facial identity, hairstyle, and clothing, while faithfully following text prompts. This enables control over expression, head pose, body gesture, and style through simple text instructions.

\subsection{Multi-Attribute Single-ID Personalization}\label{ssec:single_id_multi_attribute}
This experiment focuses on a single-subject version of \cref{ssec:multi_attr_multi_id}. We compare our method with OmniGen~\cite{OmniGen}, which, to the best of our knowledge, is the only open-sourced baseline capable of integrating up to three visual attributes for controllable human image generation. We also include GPT-4o~\cite{ChatGPT4o} as an additional multi-modal generation reference.

As shown in \cref{fig:result:multi_attribute}, \methodname~achieves significantly better preservation of identity, hair, and clothing compared to both OmniGen and GPT-4o. OmniGen often fails to follow text prompts accurately, while GPT-4o struggles to compose multiple attributes from different sources and does not preserve identity effectively in the single-subject setting. In contrast, \methodname~successfully preserves distinct attributes, integrates them naturally, and faithfully follows the text prompts.

\cref{tab:numbers} shows the quantitative results of ours compared to the state-of-the-art. From the metrics, our \methodname~achieves the highest identity perservation with similar level of image quality. 

\begin{figure}[t]
\centering
\includegraphics[width=1.0\columnwidth]{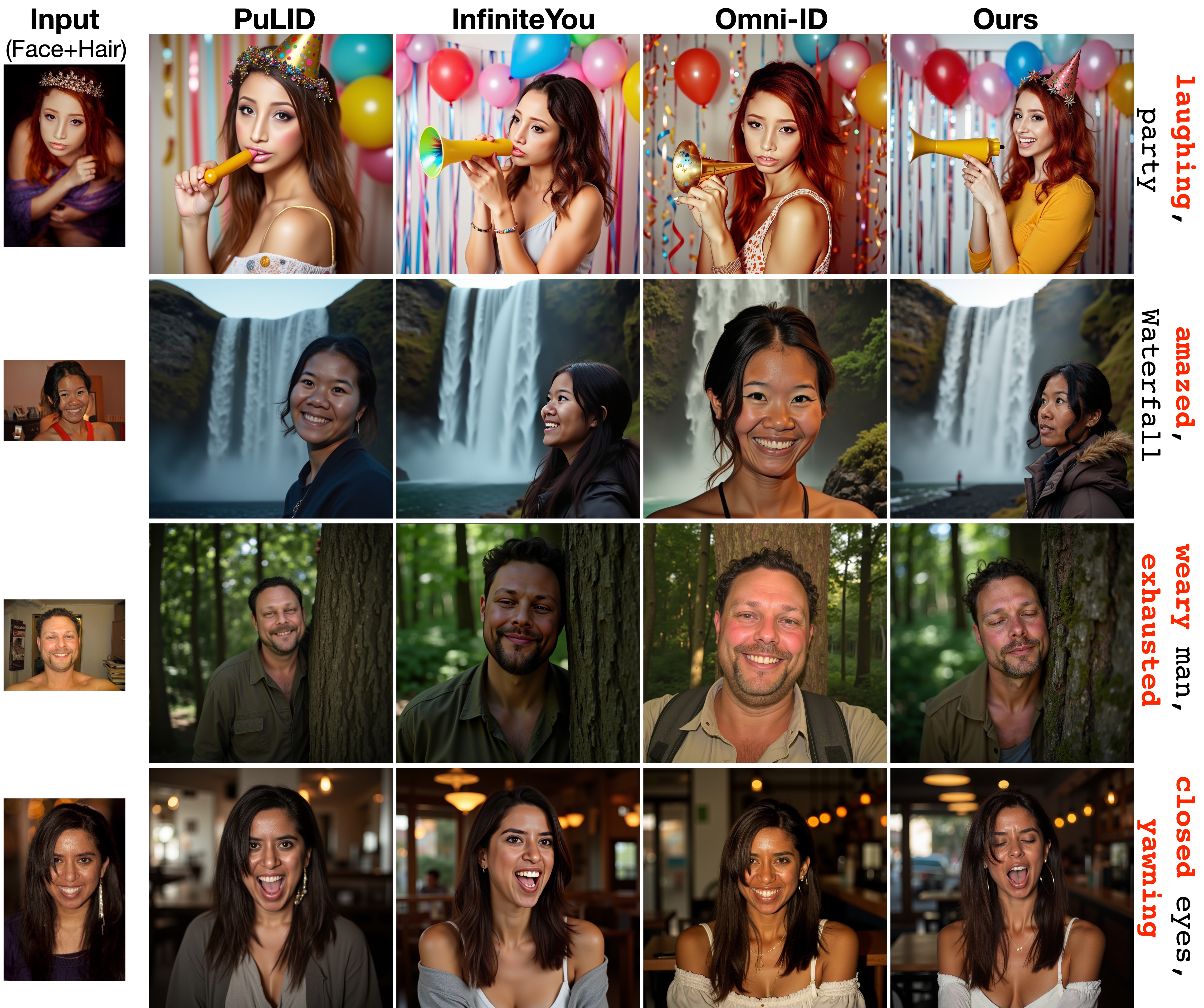}
\caption{\textbf{Qualitative Comparisons in Single-Attribute, Single-ID Personalization.} We compare \methodname~with recent PuLID, InfiniteYou, and Omni-ID, all built upon the \texttt{Flux dev} backbone. Ours demonstrates superior expressiveness, accurately following the head pose and expression specified in the text prompt while faithfully preserving subject identity.} 
\label{fig:result:single_attribute}
\end{figure}

\begin{figure}[t]
\centering
\includegraphics[width=1.0\columnwidth]{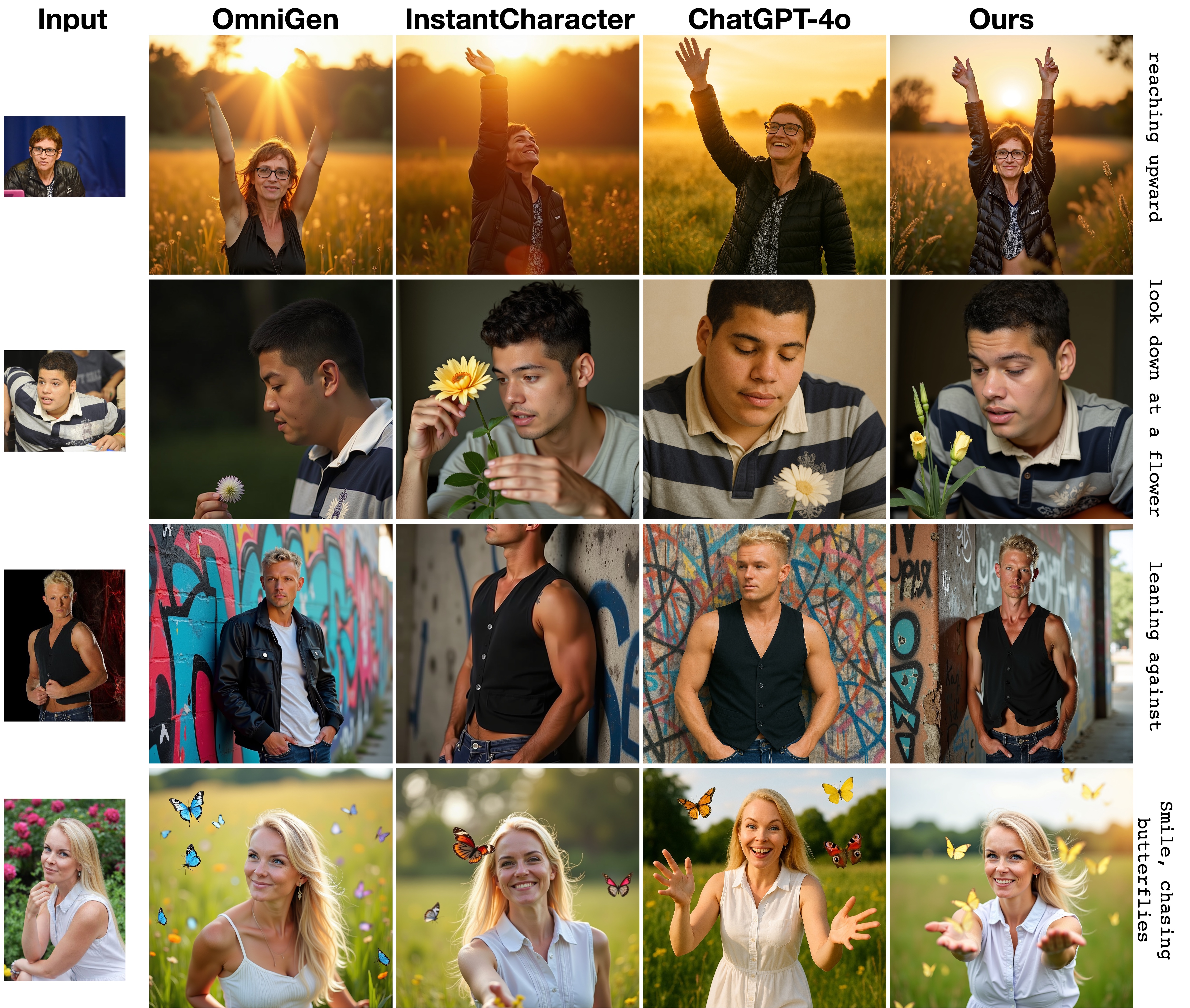}
\caption{\textbf{Qualitative Comparisons in Single-ID Full-Body Personalization.} Compared to the  state-of-the-art methods, our \methodname~again achieves superior preservation of ID, hair, and clothing attributes.
} 
\label{fig:result:fullbody_1p}
\end{figure}

\begin{figure*}[t]
\centering
\includegraphics[width=0.94\textwidth]{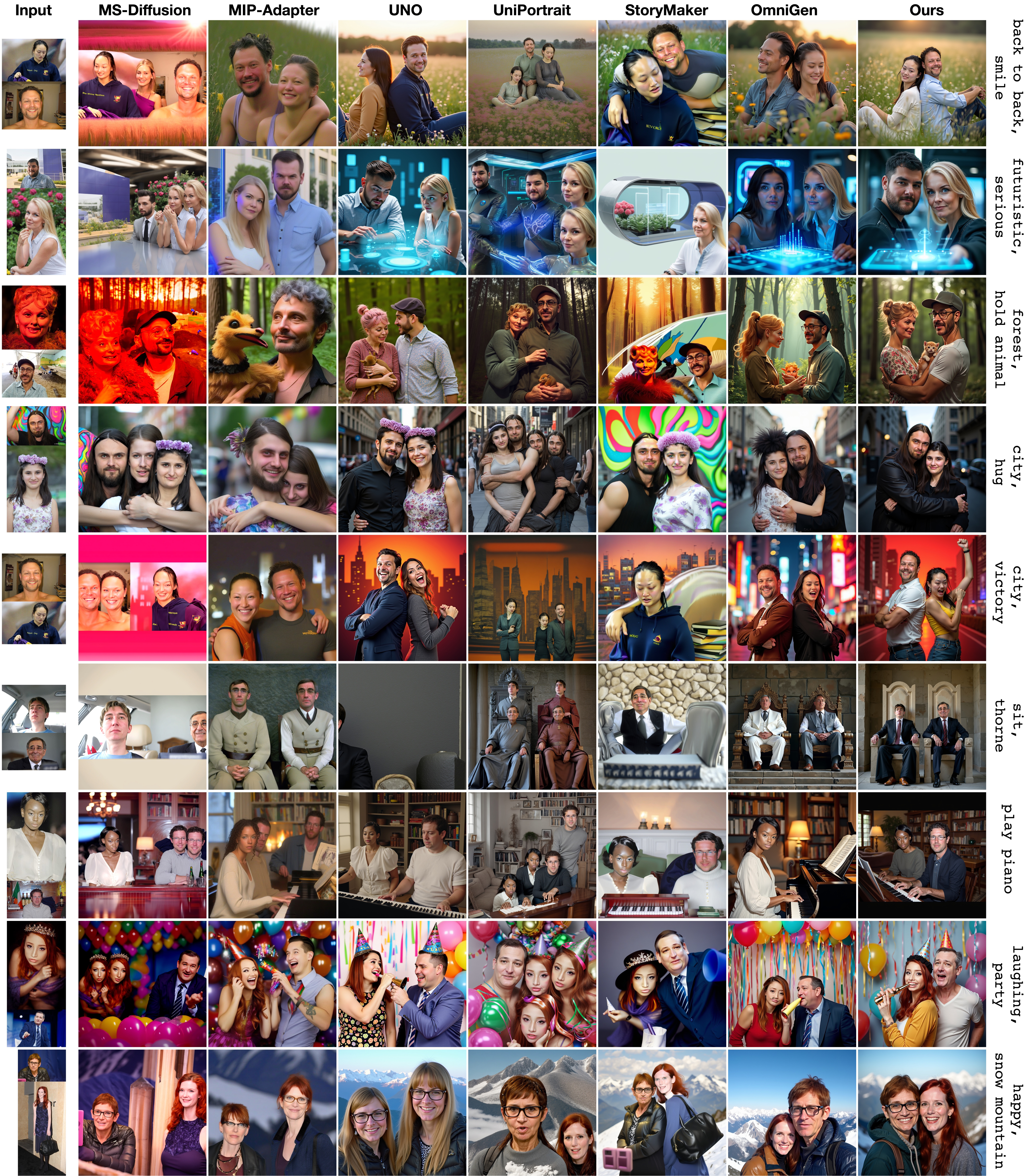}
\caption{\newtext{\textbf{Qualitative Comparisons in Single-Attribute, Multi-ID Personalization.}} \newtext{Compared to the state-of-the-art~\cite{MS-Diffusion, MIP-Adapter, UNO, he2024uniportrait, OmniGen}, \methodname~demonstrates the highest identity preservation along with the highest composition quality and the highest level of overall image quality.}} 
\label{fig:result:single_attribute_2id}
\end{figure*}

\subsection{Single-Attribute Single-ID Personalization}
The majority of prior works in personalized text-to-image generation are designed for single-attribute, single-ID personalization, where a single image is used to define one subject for customizing the generation. 
Although \methodname~is designed for multi-attribute inputs, it also seamlessly handles single-attribute personalization by using the same image for all attributes and substituting missing attributes with black images. In this experiment, both the face and the hairstyle are derived from the same input image, consistent with other methods, while a black image is used as the clothing input for our approach.

We compare our method with the most recent works in the literature, including PuLID~\cite{PuLID}, InfiniteYou~\cite{InfiniteYou}, and Omni-ID~\cite{Omni-ID}.
As shown in \cref{fig:result:single_attribute}, \methodname~qualitatively outperforms previous approaches in terms of prompt adherence. While prior methods tend to replicate the same expression and head pose from the reference image, our method generates high-fidelity results that follow the expression and head pose specified in the prompt, while still preserving identity to the highest level compared to the state-of-the-art approach~\cite{Omni-ID}.
Quantitative results are presented in \cref{tab:numbers}. Our \methodname~outperforms the state-of-the-art in terms of human preference score (HPSv2~\cite{wu2023human} and CLIP distance between the generated images and the text prompt, indicating a better prompt following. 

\cref{fig:result:fullbody_1p} further compares our method with recent works such as InstantCharacter~\cite{tao2025instantcharacter} and OmniGen, and includes GPT-4o as a reference, in the context of full-body single-ID generation. In this setting, the face, hairstyle, and clothing are all sourced from the same single image per example, i.e. the entire subject including clothing, is treated as a single attribute.
As shown, \methodname~clearly preserves identity, hairstyle, and clothing, whereas other methods struggle to maintain consistency across these visual attributes.


\subsection{Single-Attribute Multi-ID Personalization}
Similar to the single-ID, single-attribute setting, we use the same image for both face and hairstyle, and substitute the clothing input with a black image, since prior methods do not support multi-attribute input. 
We compare ours with the most recent state-of-the-art methods MIP-Adapter~\cite{MIP-Adapter}, MS-Diffusion~\cite{MS-Diffusion}, UNO~\cite{UNO}, UniPortait~\cite{he2024uniportrait}, StoryMaker~\cite{StoryMaker}, and OmniGen~\cite{OmniGen} in \cref{fig:result:single_attribute_2id}. As shown, our method achieves the highest identity preservation along with much stronger composition and much better overall image quality.

Our method also supports multi-ID full-body personalization, which is similar to \cref{fig:gallery} but the face, hair, and clothing are from the same input image. The most relevant baselines are OmniGen and GPT-4o; however, as demonstrated in previous settings, these methods fall short in preserving identity, hairstyle, and clothing with the same level of fidelity as our approach as shown in \cref{ssec:multi_attr_multi_id}. Due to space constraints and the lack of prior methods capable of producing comparable high-quality results in this setting, we omit direct comparisons.

\subsection{Ablation Study}
\begin{figure*}[t]
\centering
\includegraphics[width=1.0\textwidth]{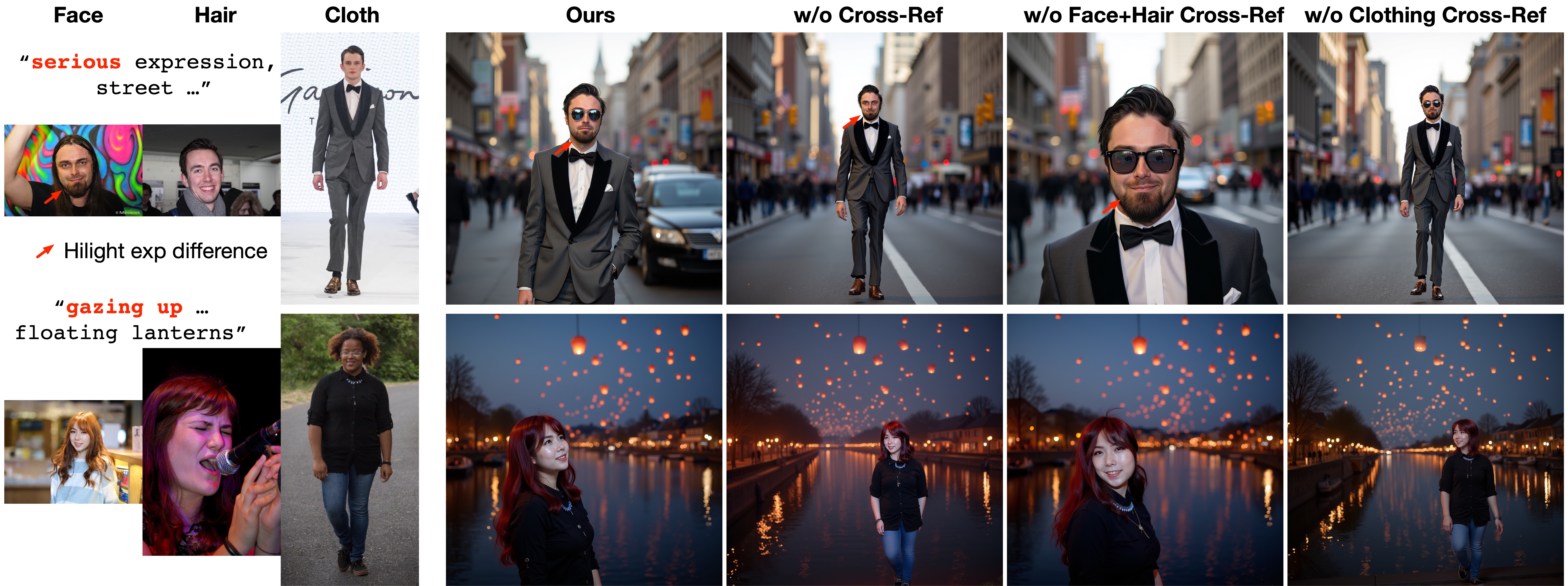}
\caption{\textbf{Qualitative Ablation Study on Multi-Attribute Cross-Reference Training.} Cross-reference training for the face and hair enables effective control over expression and head pose, while cross-reference training for clothing mitigates pose leakage originating from clothing region. When applied across all parts, multi-attribute cross-reference training allows \methodname~to achieve high-fidelity generation from misaligned attribute-specific visual prompts.}
\label{fig:ablation:cross_ref}
\end{figure*}

\begin{figure*}[t]
\centering
\includegraphics[width=1.0\textwidth]{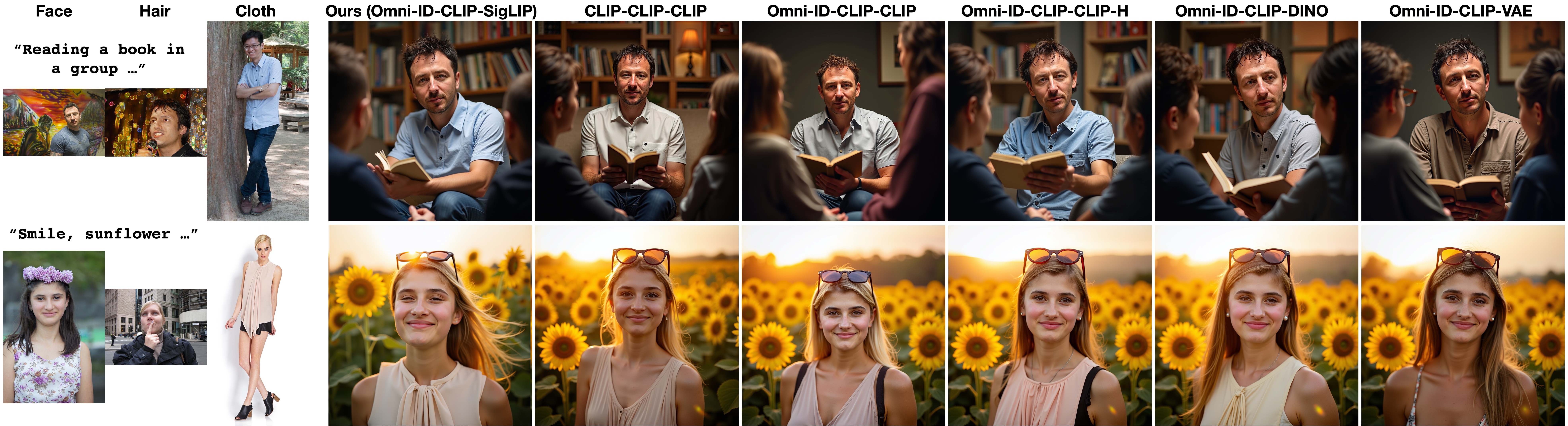}
\caption{\textbf{Qualitative Ablation Study on Attribute-Specific Tokenization.} Using Omni-ID~\cite{Omni-ID} for face, CLIP-L~\cite{CLIP} for hair, and SigLip-L~\cite{SigLIP} for clothing yields the most satisfactory results and outperforms alternative tokenization configurations.}
\label{fig:ablation:encoder}
\end{figure*}

In this section, we conduct ablation studies to analyze the impact of key components in our proposed method: the multi-attribute cross-reference training and the choice of attribute-specific tokenization.

\subsubsection{Multi-Attribute Cross-Reference Training}

\cref{fig:ablation:cross_ref} presents qualitative comparisons illustrating the impact of our proposed multi-attribute cross-reference training. As shown, our full model (``Ours'') demonstrates superior control over facial expression, head pose, and clothing details, aligning closely with both the visual and text prompts. For instance, in the bottom example, the generated image exhibits a more harmonized pose, with the head orientation accurately following the textual description.

Specifically, removing cross-reference training for face and hair (``w/o Face+Hair Cross-Ref'') leads to degraded text alignment in terms of expression and head pose. Similarly, omitting cross-reference training for clothing (``w/o Clothing Cross-Ref'') introduces artifacts, such as pose leakage from the clothing reference image, resulting in unnatural or misaligned generations, as visible from both examples.

These results underscore that multi-attribute cross-reference training is vital for achieving high-fidelity image generation. It enables precise, disentangled control over face, hair, and clothing by effectively leveraging attribute-specific visual prompts, even when these prompts exhibit misalignments. It is also interesting to see that after first stage training, all inputs are copy-pasted in the generation, providing high attribute preservation but unnatural composition, which are then fixed by our multi-attribute cross-reference training.

\subsubsection{Attribute-Specific Tokenization}
\label{subsec:ablation_encoders}
We conduct a study on the impact of using different tokenization, shown in \cref{fig:ablation:encoder}. Our proposed design, denoted as “Ours (Omni-ID–CLIP–SigLIP)”, utilizes Omni-ID~\cite{Omni-ID} for facial tokens, CLIP-L~\cite{CLIP} for hairstyle, and SigLIP-L~\cite{SigLIP} for clothing. We compare this configuration against several alternatives:
(1) CLIP–CLIP–CLIP,
(2) Omni-ID–CLIP–CLIP,
(3) Omni-ID–CLIP–CLIP-H,
(4) Omni-ID–CLIP–DINO,
(5) Omni-ID–CLIP–VAE,
where CLIP-H refers to the CLIP huge variant~\cite{CLIP}, and VAE refers to FLUX.1-dev’s VAE.

The uniform CLIP–CLIP–CLIP baseline, for instance, fails to capture the distinct visual nuances of each attribute, leading to weaker identity preservation and less detailed clothing. Our final configuration—Omni-ID (enhanced identity preservation), CLIP-L (effective hair representation), and SigLIP-L (enhanced clothing detail)—consistently outperforms all alternatives. This highlights the advantage of using specialized tokenizers for different attributes, resulting in superior overall fidelity and attribute consistency.
This study highlights the importance of attribute-specific image prompts and tokenization to achieve optimal performance.

\subsection{Implementation Details} 
\subsubsection{Pipeline Details}
\newtext{We experiment \methodname~with $K = 3$, corresponding to three distinct visual attributes: face, hairstyle, and clothing. \texttt{FLUX.1-dev}~\cite{FLUX} is employed as the base model and remains frozen during training. Decoupled attentions are learned and applied only within the single-stream blocks (i.e., block indices 19 to 56), as this is found to be sufficient. The hyperparameter $\lambda$ is set to 1, and the number of denoising steps is fixed at 28 across all experiments.}

\begin{figure*}[t]
  \includegraphics[width=\textwidth]{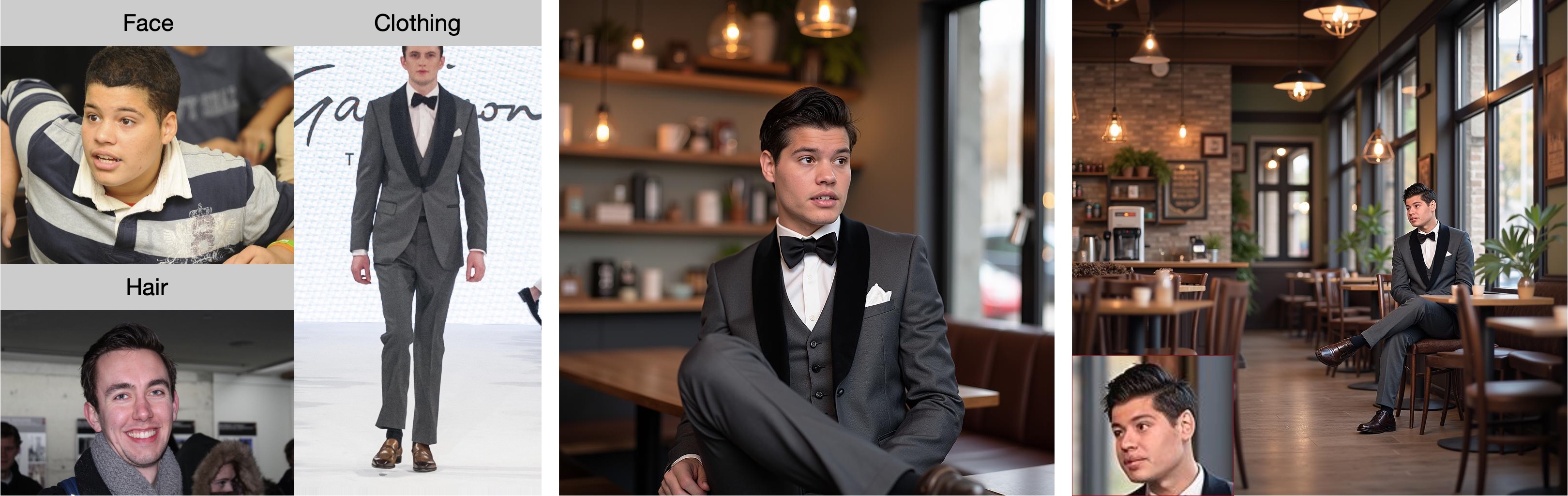}
  \caption{\newtext{Failure case. Given the input face, hair, and clothing sources (left), \methodname{} produces high-quality images when the generated face is sufficiently large. However, when the text prompt specifies a distant shot, the face becomes small, leading to noticeable distortions (right). This is caused by a training-data filter that removed small-face samples; including them could alleviate the issue.}}
  \label{fig:fail}
\end{figure*}

\subsubsection{Training Details} \newtext{We train decoupled attentions on a frozen \texttt{FLUX.1-dev} using AdamW with a constant lr 1e-4, a total batch size of 8, and image resolution of $512\times512$, for 200k iterations per stage on 8×A100s (4 GPU-days per stage). 
Note \texttt{FLUX.1-dev} is a guidance distilled finetuned on highly aesthetic images. To capture the distillation artifacts as well as the distribution shift of our training set regards to the base model, we train a LoRA with rank $512$ over our training dataset with guidance scale equal to $1$. This LoRA is frozen during our adapter training, and removed in inference time. 
Our training set consists of around 32M in-house images covering 6M identities, annotated by internal instance/clothing/face segmentation models. The training set is filtered to include up to 2 identities.}

\subsubsection{Evaluation Details} 
\newtext{We evaluate at $1024$ resolution using 32 diverse prompts, the first 32 images from FFHQ-in-the-wild~\cite{StyleGAN} as faces, and 64 random SSHQ~\cite{StyleGAN-Human} images as hair/clothing. Both are public single-frame datasets not included in training. Across all evaluations, $\lambda=1$ and denoising steps $28$ are used. We do not use any per-prompt tuning or post processing.} 

\newtext{In terms of metrics, for identity preservation, we use FaceNet~\cite{schroff2015facenet}.
For hair preservation, we align the cropped face to $1.4\times$ the size used in FFHQ~\cite{StyleGAN}, then segment the hair using FaRL~\cite{FaRL}.
CLIP features are extracted from the resized, segmented hair and used to measure the similarity between the generation and the source.
For clothing preservation, we apply our clothing segmentation model, crop the clothing to its bounding box, and then center-crop the segmented clothing. CLIP features are computed on the resulting cropped clothing image.
In multi-subject cases, for each attribute in the source, we match it to the most similar generated attribute by finding the closest distance.}

\subsubsection{Baseline Details}

\noindent\newtext{\textbf{OmniGen.} In multi-ID, single-attribute personalization, we use the following OmniGen instruction:
\texttt{"\{prompt\}. The first person is <img><|image\_1|></img> and the second person is <img><|image\_2|></img>."}
In single-ID, multi-attribute personalization, the instruction is:
\texttt{"\{prompt\}. This person is <img><|image\_1|></img>, with hair <img><|image\_2|></img> and wear <img><|image\_3|></img>."}
where \{\texttt{prompt}\} is the same as \methodname. The inference for each example of OmniGen takes $50$ denoising steps}.

\noindent\newtext{\textbf{GPT4o.} 
\newtext{We use the following GPT4o instruction}: \texttt{"Make the person in the first image with the hairstyle from the second image and wear clothes from the third image. Generate image using this person following this prompt: \{prompt\}"}.}

\newtext{For all other baselines in the paper, we follow the official code or paper without changing the default parameters.}

\section{Limitations and Failure Cases}
\noindent\newtext{\textbf{Limitations.} ComposeMe is a closed-set personalization method trained for up to 2 identities with 3 attributes. For most human-centric use cases—face, hair, clothing, and pose—this coverage is sufficient. Pose control already works natively via ControlNet, similar to all other IP-Adapters. To extend ComposeMe with new visual prompts (e.g., scene, accessories), we identify two paths: (1) finetuning ComposeMe by concatenating new visual prompts, and (2) training new adapters for additional categories and combining them with ComposeMe at inference. The latter also supports fine-tuning for improved quality. In addition, from metrics and general image quality, we do see the gap between ComposeMe and GPT-4o. Finetuning on high-quality data and RL tuning offer promising directions in improving the quality. We leave these as future work.} 

\noindent\newtext{\textbf{Failure cases.}
The primary failure case occurs when the generated faces are tiny (<1\% of image), resulting in noticeable face distortions (see \cref{fig:fail}. We traced this to a training-data filter that removed small-face samples. Including small faces in the training is a reasonable solution to address these failures. }

\section{Conclusion}
\methodname{} introduces a novel and effective approach for fine-grained, controllable human image generation. By the proposed Attribute-Specific Image Prompts, our method enables composable synthesis from distinct visual sources. Multi-Attribute Cross-Reference Training is introduced for achieving robust disentanglement and natural composition. This strategy allows the model to generate harmonized outputs from misaligned attribute inputs while faithfully adhering to textual conditioning.

Extensive experiments demonstrate that \methodname{} achieves state-of-the-art performance, effectively following both visual and textual prompts. 
\methodname{} paves the way for controllable and modular human image generation, offering a powerful and extensible paradigm for future personalization applications.

\begin{acks}
The authors would like to acknowledge Ke Ma and Huseyin Coskun for infrastructure support; 
Ruihang Zhang, Or Patashnik and Daniel Cohen-Or for their feedback on the paper; the anonymous reviewers for their constructive comments;
and other members of the Snap Creative Vision team for their valuable feedback.
\end{acks}

\bibliographystyle{ACM-Reference-Format}
\bibliography{main}

\end{document}